\documentclass[lettersize,journal]{IEEEtran}
\usepackage{amsmath,amsfonts}
\usepackage{algorithmic}
\usepackage{algorithm}
\usepackage{array}
\usepackage[caption=false,font=normalsize,labelfont=sf,textfont=sf]{subfig}
\usepackage{textcomp}
\usepackage{stfloats}
\usepackage{url}
\usepackage{verbatim}
\usepackage{graphicx}
\usepackage{tabularray}
\usepackage{cite}
\usepackage{lipsum}
\usepackage{tabularx}
\usepackage{multirow}
\usepackage{makecell}

\begin{document}

\title{Integrated Dynamic Phenological Feature for Remote Sensing Image Land Cover Change Detection}

\author{Yi Liu,~\IEEEmembership{member,~IEEE,} Chenhao Sun, Hao Ye, Xiangying Liu and Weilong Ju
\thanks{This paper was produced by the IEEE Publication Technology Group. They are in Piscataway, NJ.}
\thanks{Manuscript received April 19, 2021; revised August 16, 2021.}}

\markboth{Journal of \LaTeX\ Class Files,~Vol.~14, No.~8, August~2021}%
{Shell \MakeLowercase{\textit{et al.}}: A Sample Article Using IEEEtran.cls for IEEE Journals}


\maketitle

\begin{abstract}
Remote sensing image change detection (CD) is essential for analyzing land surface changes over time, with a significant challenge being the differentiation of actual changes from complex scenes while filtering out pseudo-changes. A primary contributor to this challenge is the intra-class dynamic changes due to phenological characteristics in natural areas. To overcome this, we introduce the InPhea model, which integrates phenological features into a remote sensing image CD framework. The model features a detector with a differential attention module for improved feature representation of change information, coupled with high-resolution feature extraction and spatial pyramid blocks to enhance performance. Additionally, a constrainer with four constraint modules and a multi-stage contrastive learning approach is employed to aid in the model's understanding of phenological characteristics. Experiments on the HRSCD, SECD, and PSCD-Wuhan datasets reveal that InPhea outperforms other models, confirming its effectiveness in addressing phenological pseudo-changes and its overall model superiority. Our code and data is available
at https://github.com/Augety-Sun/InPhea.
\end{abstract}


\section{Introduction}
\IEEEPARstart{C}{hange} Detection (CD) in remote sensing (RS) is a popular research topic that involves capturing surface changes using bi-temporal or multi-temporal remote sensing image data\cite{ref1}. It finds extensive applications in urban management, disaster assessment, and environmental monitoring. 

Currently, with the diversification of data sources, application scenarios, and target features, remote sensing change detection tasks face a series of challenges, primarily including (1) precise identification of target areas in complex scenes and (2) filtering of 'pseudo-changes' at the semantic level within classes.

For the first challenge, the research and development of deep learning models provide a promising approach to address this issue. Some scholars have made preliminary progress by using Siamese neural networks for change detection \cite{ref2,ref3,ref4}, but they did not consider the semantic information of land features, resulting in limitations in detecting changes in complex scenes. Other researchers have successfully improved the accuracy of change detection by introducing dual-branch networks\cite{ref5,ref6}. Despite the improvement in accuracy brought about by the incorporation of segmentation, the current predominant approach of utilizing small-batch learning in training poses challenges in mitigating inter-class similarity interference, particularly in complex scenarios\cite{ref7}. Consequently, this paper proposes the application of contrastive learning alongside change detection to enhance inter-class discrimination, thereby addressing the challenges presented by complex scene variations.

For the second challenge, the key reason for the formation of 'pseudo-changes' at the semantic level within classes is that phenological features of natural areas cause seasonal variations in similar land features, which are not the changes we truly want to identify and need to filter out. Hence, it is necessary to study surface phenological features and explore methods that enable models to understand and filter surface phenological changes. Yang et al. \cite{ref8} constructed a rice growth stage recognition network based on the AlexNet architecture, demonstrating outstanding performance of CNNs in phenological detection. Yalcin et al. \cite{ref9} built a classification network based on AlexNet to classify different phenological stages of six plants, proving the effectiveness of CNN-based methods in phenological classification. Song et al. \cite{ref10} developed a comprehensive database covering long-term leaf characteristics and utilized a superpixel-based residual network (SP-ResNet50) to automatically differentiate between leaves and non-leaves in images, demonstrating the network's ability to help distinguish leaf phenological characteristics. The aforementioned studies show that convolutional neural networks perform well in the study of surface phenology, as they can learn and understand phenological features. However, they typically learn single categories and do not simultaneously consider multiple classes, which makes them prone to failure in complex and large-scale region. Therefore, this paper focuses on extracting and learning phenological features of multiple coexisting land cover classes in complex scenes.

Based on the aforementioned issues and needs, we propose the InPhea method for change detection utilizing joint phenological features, providing a solution to mitigate the influence of natural phenology and accurately identify changes in complex environments. The main contributions of this work can be summarized as follows:

\begin{enumerate}
\item{We introduce a detector-constrainer architecture for change detection and employ a multi-stage contrastive learning strategy, using the constrainer to guide the learning of the detector gradually. Additionally, we propose a differential attention module and four constraint modules targeting phenological issues to enhance the detection performance of the model.}
\item{We curate a change detection remote sensing dataset, PSCD-Wuhan, specifically designed for addressing surface phenology issues. This dataset comprises samples with multiple classes, long time-series, and various phenological features, providing abundant phenological information for model training while alleviating the scarcity of remote sensing data in the industry focusing on phenological problems.}
\item{Compared to other state-of-the-art change detection methods, InPhea demonstrates superior performance on the PSCD-Wuhan, HRSCD, and SECD datasets.}
\end{enumerate}

The remainder of this paper is organized as follows: Section II describes our methodology. Section III presents the experimental results and analysis. Finally, Section IV concludes our study.

\begin{figure*}[ht]
\centering
\includegraphics[width=\textwidth]{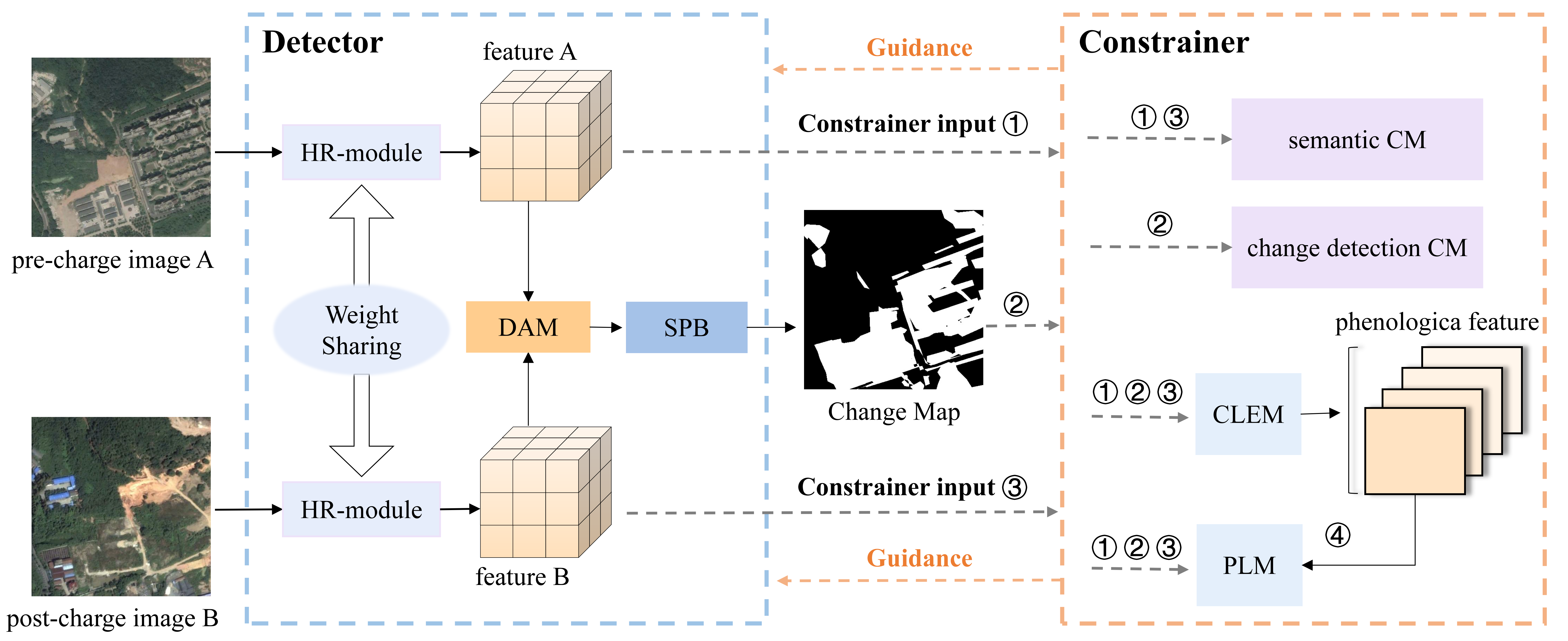}
\caption{The overall architecture of InPhea.}
\label{fig:wide-image}
\end{figure*}

\section{Method}
We introduce a remote sensing image change detection method, termed InPhea, leveraging joint phenological features. As illustrated in Fig. 1, it comprises two components: the detector and the constrainer. In the detector component, we devise a feedforward change detection backbone network. Initially, the network employs a twin-structured high-resolution feature extraction module (HR Module, employing HR net18) to extract features from dual-temporal images. Subsequently, the extracted features are fed into a difference attention module (DAM) to fuse and locate the differential information between the dual-temporal images. The differential information is then processed through a spatial pyramid block (SPB) for multi-scale feature extraction and fusion, ultimately entering a pre-trained segmentation module (utilizing FCN) to obtain change detection results. The constrainer component comprises four modules: Change Detection CM guides the network to focus on change information, Semantic CM introduces prior semantic information into the network, Contrastive Learning Phenological Feature Extraction Module (CLEM) addresses inter-class similarities, promoting the aggregation of features of the same class to foster the model's phenological feature understanding capability, and Phenological Learning Module (PLM) tackles intra-class diversity, further enhancing the model's phenological feature understanding capability.

During the prediction phase, only the Detector is utilized in the change detection process. However, during the training phase, the Constrainer is employed to guide the Detector. In multiple stages, the Constrainer's aforementioned modules are utilized to calculate LOSS and backpropagate to update parameters, guiding the Detector's training in four distinct aspects, enabling it to focus on the required changes. Below are some details of the method.

\subsection{Detector}
\subsubsection*{\bf 1) Differential Attention Module}This module integrates various differential information extraction methods, fully considering the spectral and positional information of the dual-temporal feature maps. It enlarges the feature differences between changing and unchanged regions, thereby improving detection accuracy. As illustrated in the Fig. 2, the feature maps are separately input into two branches for processing. One branch utilizes feature concatenation and convolution to obtain differential features, while the other parallel branch calculates attention feature maps considering both channel and positional information to enhance differential information. Finally, the results from both branches are multiplied to obtain the change feature map based on differential attention.

\begin{figure}[h]
    \centering
    \includegraphics[width=1\linewidth]{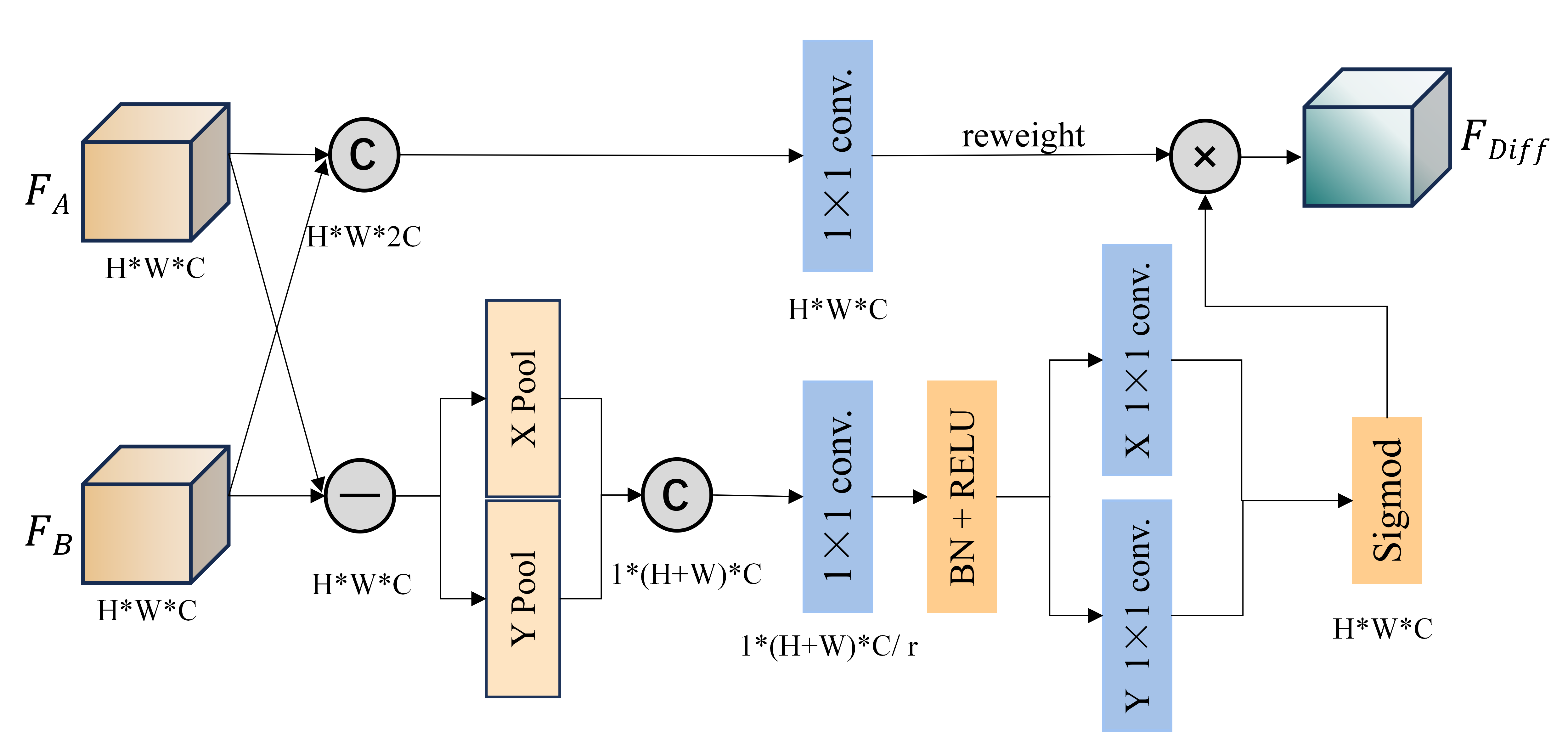}
    \caption{Differential Attention Module Diagram.}
    \label{fig:enter-label}
\end{figure}

\subsubsection*{\bf 2) Spatial Pyramid Block}For the obtained change features, we constructed a spatial pyramid block to integrate the change features at multiple scales, enhancing the model's robustness. As shown in Fig. 3, for a single-layer pyramid structure, the differential information undergoes average pooling, followed by processing through convolutional layers, BN layers, and activation function layers. Finally, it is upsampled to match the size of the original feature map and concatenated with the original feature map to obtain a multi-scale representation of the change features. This representation can be directly input into the pretrained segmentation module to obtain the change detection map.

\begin{figure}[h]
    \centering
    \includegraphics[width=1\linewidth]{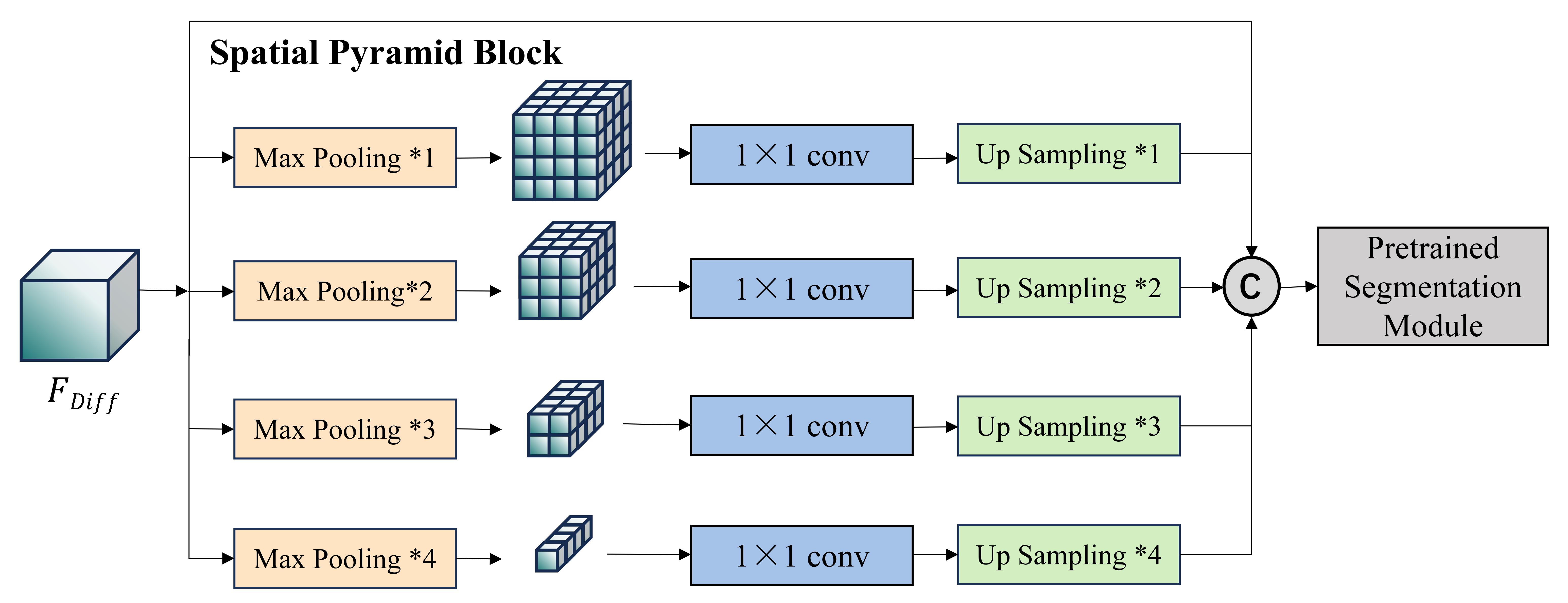}
    \caption{Spatial Pyramid Block Structure Diagram.}
    \label{fig:enter-label}
\end{figure}

\subsection{Constrainer}
\subsubsection*{\bf 1) Change Detection Constraint Module}The role of this module is to guide the detector to focus on the change information in the images. It utilizes the binary cross-entropy function to calculate the loss between the predicted change regions and the ground truth change regions, encouraging the detector to prioritize the identification of change areas.

\subsubsection*{\bf 2) Semantic Constraint Module} This module introduces prior semantic information to the detector by training a semantic segmentation task-constrained network. Leveraging the dual-temporal image features extracted by the detector, this module incorporates a semantic segmentation module (utilizing FCN) to obtain semantic maps. Subsequently, it calculates the loss between the predicted semantic map and the ground truth semantic map using cross-entropy loss, and updates the parameters of the feature extraction module in the detector via backpropagation.

\subsubsection*{\bf 3) Contrastive Learning Phenological Feature Extraction Module} This module serves two main purposes: (1) addressing similarities between different types of land features to facilitate the aggregation of characteristics within the same land type and (2) providing phenological features required for training the phenological learning module, enhancing the network's understanding and utilization of phenological information in the data. This module comprises phenological feature extraction blocks and phenological learning contrastive loss blocks. As shown in Fig. 4, the phenological feature extraction block remaps features extracted by the detector to update features and discretize representations of different classes. The phenological learning contrastive loss block first selects pixel points based on a hard negative sampling strategy (TABLE I). For selected samples, a random selection of N samples participates in contrastive learning, computing the dual-task contrastive loss for semantic segmentation and change detection tasks (Equations 1-4) to constrain network training. Here, LP-P represents pixel-to-pixel contrastive loss, LR-R denotes region-to-region contrastive loss, LP-R signifies pixel-to-region contrastive loss, and LMIX denotes the combined loss function

{\small
\begin{align}
\label{eq:loss_function}
L_{P-P} = - \mathbb{E}_{X} \Biggl[ \log \Biggl( \frac{\exp \left( f(x_h)^T f(x^+) \right)}{\exp \left( f(x_h)^T f(x^+) \right) + \sum_{j=1}^{N-1} \exp \left( f(x_h)^T f(x_j) \right)} \Biggr) \Biggr]
\end{align}
}

{\small
\begin{align}
L_{R-R} = \frac{1}{n_j} \sum_{i=1}^{n_j} x_i \quad \in (0,m)
\end{align}
}

{\small
\begin{align}
L_{P-R} = -E_X \left[ \log \left( \frac{\exp(f(x_h)^T f(x^+))}{\exp(f(x_h)^T f(x^+)) + \sum_{j=1}^{N-m-2} \exp(f(x_h)^T f(x_j))} \right) \right]
\end{align}
}

\begin{equation}
{\small
\begin{aligned}
L_{MIX} = & -E_{X_{L_{R-R}+M}} \\
& \hspace{-\leftmargin} \left[ \log \left( \frac{\exp(f(x_h)^T f(x^+))}{\exp(f(x_h)^T f(x^+)) + \sum_{j=1}^{N+m+M-2} \exp(f(x_h)^T f(x_j))} \right) \right]
\end{aligned}
}
\end{equation}

\begin{figure}[h]
    \centering
    \includegraphics[width=1\linewidth]{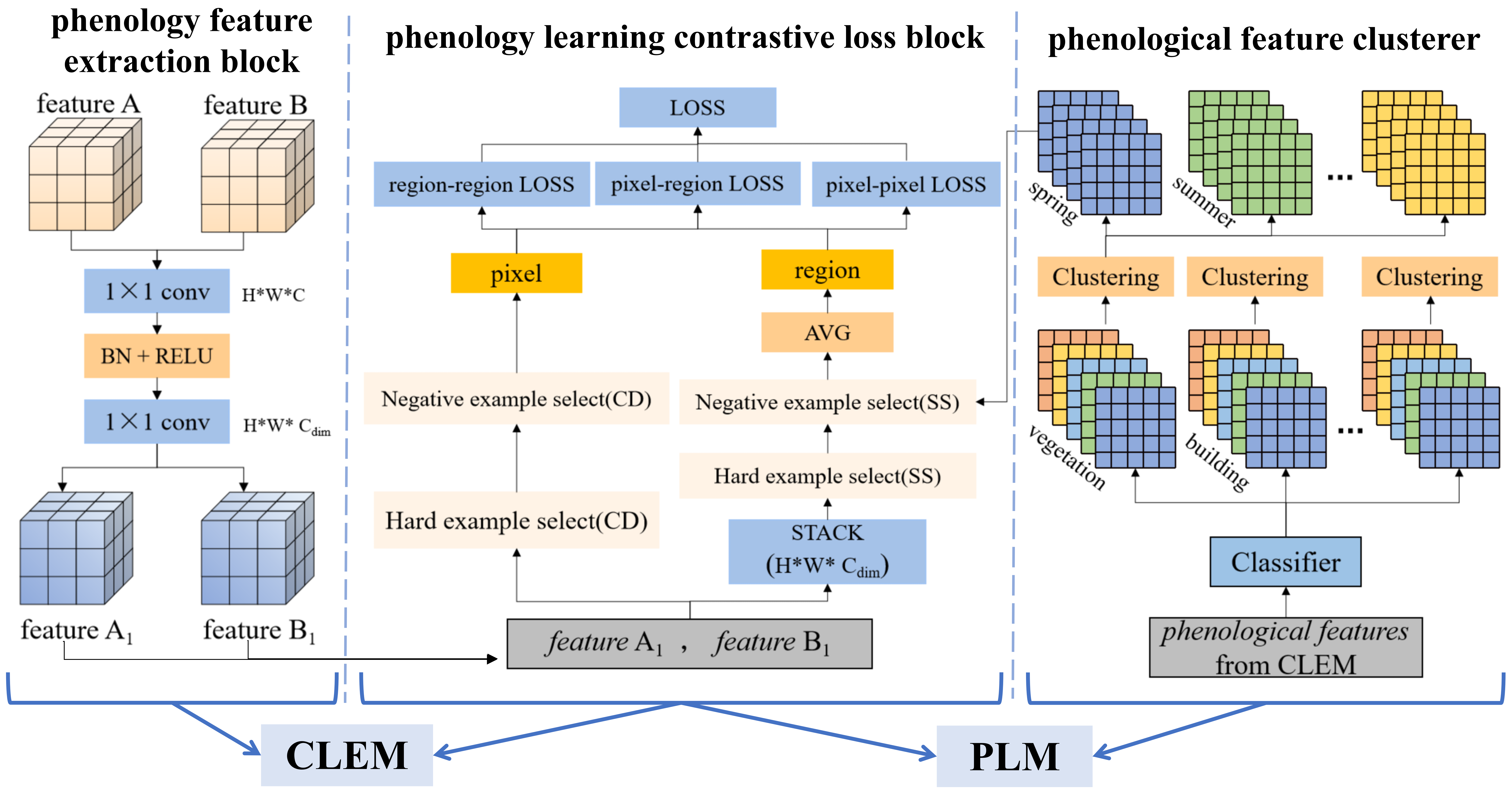}
    \caption{CLEM and PLM Structure Diagram.}
    \label{fig:enter-label}
\end{figure}

\begin{table*}[!t]
\caption{Positive and negative sample selection strategy\label{tab:table1}}
\centering
\resizebox{\textwidth}{!}{%
\begin{tblr}{
  colspec={|X|X|X|X|X|},
  row{1} = {font=\bfseries},
  hlines, vlines,
  cell{1}{1} = {c=2}{},
  cell{2}{1} = {c=2,r=2}{},
  cell{2}{3} = {r=2}{},
  cell{2}{5} = {r=2}{},
  cell{4}{1} = {r=4}{},
  cell{4}{2} = {r=2}{},
  cell{4}{3} = {r=2}{},
  cell{4}{5} = {r=2}{},
  cell{6}{2} = {r=2}{},
  cell{6}{3} = {r=2}{},
  cell{6}{5} = {r=2}{},
}
Difficult sample selection strategy &                  & Anchor sample     & Positive sample               & Negative sample \\
\hline
Segmentation                        &                  & Current point     & Similar points (CLEM)         & Non-similar points \\
\hline
                                    &                  &                   & Same phenological stage (PLM) &                   \\
\hline
Change detection                    & changed          & Current point (Category A) & Category A points (CLEM)      & Non-category A points \\
\hline
                                    &                  &                   & Same phenological stage (PLM) &                   \\
\hline
                                    & unchanged        & Current point (Category A) & Category A points (CLEM)      & Non-category A points \\
\hline
                                    &                  &                   & Same phenological stage (PLM) &                   
\end{tblr}
}
\end{table*}

\subsubsection*{\bf 4) Phenological Learning Module} This module aims to utilize the features extracted by CLEM from the training set to address intra-class diversity, enhancing the detector's phenological understanding. As depicted in Fig. 4, this module comprises a phenological feature clusterer and a phenological learning contrastive loss block. The phenological feature clusterer employs a multi-center phenological information modeling approach\cite{ref11}  to process phenological features extracted by CLEM, obtaining features from different stages within each class. The phenological learning contrastive loss block employs a contrastive learning strategy to insert single-stage phenological features into supervised contrastive learning for change detection, building upon the structure of the feature extraction contrastive loss block. This transition enables the network to discern features not only among samples of the same class but also among different phenological stages within the same class, thereby addressing the challenge of intra-class diversity comprehension. It is worth noting that, as shown in TABLE I, in this module, the phenology-aware contrastive loss block targets each anchor sample. Negative samples are selected from instances of different classes, with each instance feature randomly selected a certain number of times. Positive samples are selected from instances of the same class and belonging to the same phenological centroid class, with each instance feature also randomly selected a certain number of times.

\subsubsection*{\bf 5) Explanation of the Constrainer's Work} As shown in Fig. 5. Constrainer-assisted Multi-stage Contrastive Learning Schematic., the process of constrainer-guided detector training is conducted in three stages: In the first stage, utilizing Change Detection CM, Semantic CM, and CLEM, the detector is trained to address inter-class similarities, aggregating features of the same class, and initially cultivating the detector's phenological understanding capability. In the second stage, leveraging the detector with the best training metrics and CLEM, phenological features of the training dataset are generated, followed by clustering of phenological features using PLM's phenological feature cluster. In the third stage, utilizing Change Detection CM, Semantic CM, CLEM, and PLM, the detector is trained. In long temporal data, intra-class diversity significantly increases, and there are substantial differences in features among different phenological stages within the same class, resulting in increased difficulty in network training. Therefore, it is essential to conduct more precise contrastive learning based on phenological stages as the basic unit to reduce network learning difficulty. During this stage, not only do features between different classes become distant, but features between the same class but different phenological stages also become distant, with features only aggregating within the same class and phenological stage.

\begin{figure}
    \centering
    \includegraphics[width=1\linewidth]{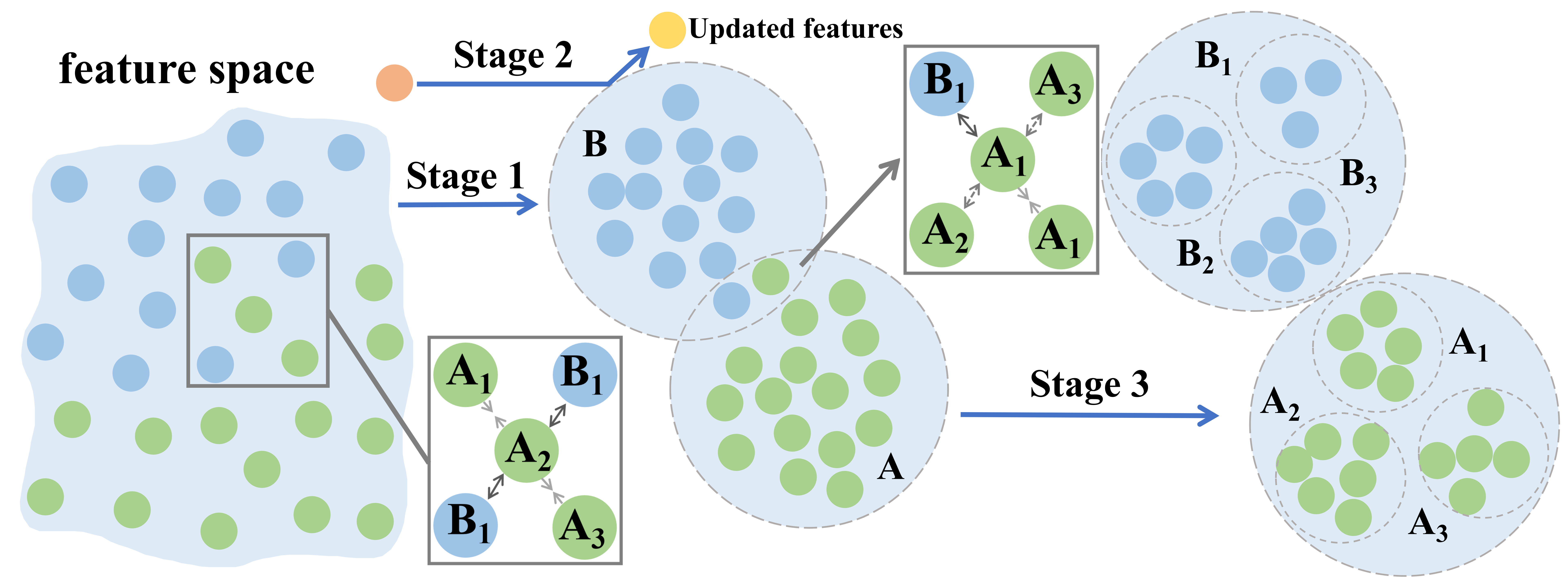}
    \caption{Constrainer-assisted Multi-stage Contrastive Learning Schematic.}
    \label{fig:enter-label}
\end{figure}

\section{Experimental Results And Analysis}
We evaluated InPhea's effectiveness using three datasets: HRSCD\cite{ref12}, SECD(Sense Earth 2020 change detection dataset), and the PSCD-Wuhan dataset. SECD, from the 2020 AI Remote Sensing Interpretation Competition by SenseTime, consists of 4662 pairs of dual-temporal images with corresponding semantic maps, all sized at 512×512 pixels. Land cover types include ground, low vegetation, trees, water bodies, buildings, sports fields, and others. HRSCD covers areas in Brittany, Loire, and Normandy regions of France, comprising 5365 pairs of dual-temporal images, each sized at 1000×1000 pixels, with land cover types including buildings, agricultural land, forests, wetlands, water bodies, and others. PSCD-Wuhan is a self-collected dataset with complete seasonal phenological information, containing 2400 data sets, each including remote sensing images, semantic maps, and change maps corresponding to a complete seasonal phenological cycle (such as spring, summer, autumn, winter), with image size at 256×256 pixels. Land cover types include 10 categories: grassland, shrubs, built-up areas, bare land, crops, trees, water bodies, paddy fields, fallow land, and others. Each dataset is split into training, validation, and testing sets in an 8:1:1 ratio.

We used PyTorch as the deep learning framework, conducting experiments on an NVIDIA GTX 3090 workstation. We employed the SGD solver with a momentum of 0.9, weight decay of 0.0001, and a base learning rate of 0.25\%. Each experiment was trained for 300 epochs, with validation performed every 5 epochs. Evaluation was based on F1 score and Intersection over Union (IoU). Evaluation was conducted based on the cumulative confusion matrix of all test images: TP represents the number of correctly predicted change samples, TN represents the number of correctly predicted unchanged samples, FN represents the number of samples incorrectly predicted as changed, and FP represents the number of samples incorrectly predicted as unchanged. Evaluation metrics were calculated according to the following formulas.

\begin{equation}
\label{eq:precision}
P = \frac{TP}{TP + FP}
\end{equation}

\begin{equation}
\label{eq:recall}
\text{Recall} = \frac{TP}{TP + FN}
\end{equation}

\begin{equation}
\label{eq:F1}
F1 = \frac{2 \times P \times \text{Recall}}{P + \text{Recall}}
\end{equation}

\begin{equation}
\label{eq:IOU}
\text{IOU} = \frac{TP}{TP + FP + FN}
\end{equation}

\begin{figure}[h]
    \centering
    \includegraphics[width=1\linewidth]{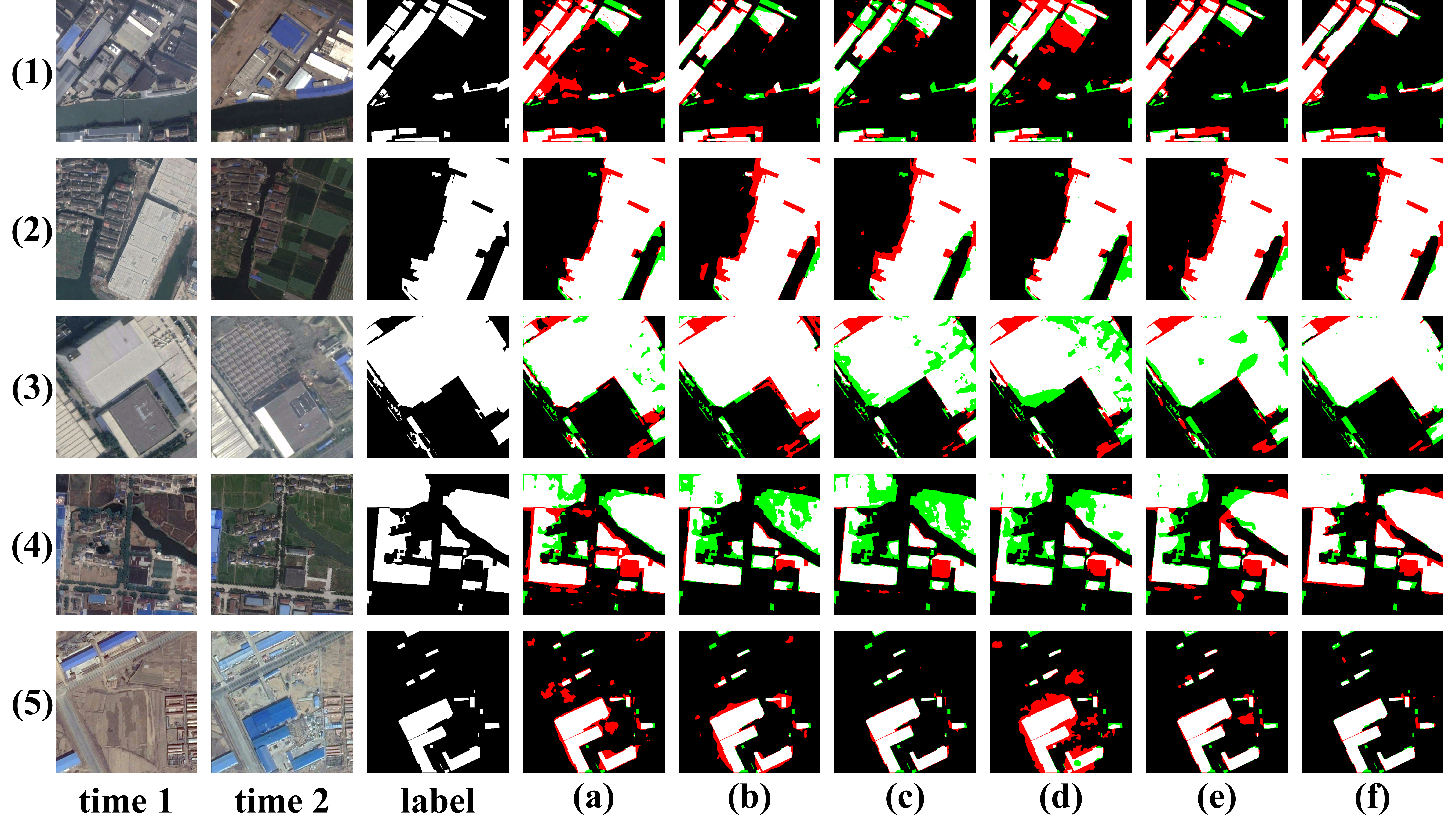}
    \caption{Visualization Results of Various Change Detection Methods.(a) FC-EF.(b) FC-Siam-conc.(c) FC-Siam-diff.(d) Siam-Deeplabv3.(e) Siam-UNet. (f) ours.}
    \label{fig:enter-label}
\end{figure}

\begin{table}[h]
\caption{COMPARISON WITH THE SOTA METHODS\label{tab:datasets}}
\centering
\begin{tabular}{|c||c|c|c|c|c|c|}
\hline
\textbf{Dataset}        & \multicolumn{2}{c|}{\textbf{SECD}}   & \multicolumn{2}{c|}{\textbf{HRSCD}}  & \multicolumn{2}{c|}{\textbf{PSCD-Wuhan}} \\
\cline{2-7}
                        & \textbf{IOU}   & \textbf{F1}    & \textbf{IOU}   & \textbf{F1}    & \textbf{IOU}        & \textbf{F1}    \\
\hline\hline
\textbf{FC-EF}          & 67.01          & 78.09          & 66.18          & 75.87          & 73.07               & 83.8           \\
\hline
\textbf{FC-Siam-conc}   & 67.23          & 78.12          & 66.02          & 75.25          & 72.58               & 83.67          \\
\hline
\textbf{FC-Siam-diff}   & 66.55          & 77.2           & 65.54          & 74.32          & 71.46               & 82.93          \\
\hline
\textbf{Siam-DeepLabv3} & 67.41          & 78.87          & 66.02          & 75.35          & 71.88               & 82.89          \\
\hline
\textbf{Siam-Unet}      & 67.97          & 79.95          & 66.51          & 75.99          & 71.69               & 83.34          \\
\hline
\textbf{Ours}           & \textbf{71.88} & \textbf{82.76} & \textbf{69.43} & \textbf{78.92} & \textbf{79.41}      & \textbf{88.25} \\
\hline
\end{tabular}
\end{table}

TABLE II presents the F1 score and IOU results of different methods on three datasets. It is evident that InPhea outperforms competing methods, with IOU scores on the three datasets surpassing those of the second-best model by 3.94\%, 2.92\%, and 6.34\%, respectively, while F1 scores exceed those of the second-best model by 2.81, 2.93, and 4.45. Fig. 6 illustrates visual comparisons of partial data from different methods across three datasets. It can be observed that InPhea effectively filters out variations caused by phenology. For instance, in images 1, 2, and 3, the model accurately identifies changing regions while filtering out pseudo-changes in water bodies with varying hues. Additionally, in images 1 and 3, the model filters out pseudo-changes occurring due to vegetation in different seasons, accurately identifying genuine changes in buildings. Furthermore, the model demonstrates increased sensitivity to small-scale changes and exhibits good overall integrity in the results.

Furthermore, we conducted ablation experiments on the components of InPhea, demonstrating their effectiveness. TABLE III shows the results of detector component ablation experiments. The introduction of the Differential Attention Module (DAM) led to an improvement of 1.78\% and 2.27\% in IOU when compared to using channel concatenation or feature map subtraction alone, and F1 scores also increased by 0.67 and 1.39, respectively. Further enhancements were observed with the inclusion of the spatial pyramid block, indicating better utilization of difference information. TABLE IV presents the results of ablation experiments on the constraint modules, showing that the addition of various constraint modules led to improvements in model accuracy across all datasets, underscoring the effectiveness of the prior information introduced to the detector through these modules.

\begin{table}[h]
\caption{Ablation Experiment Results of the Detector on the SECD Dataset\label{tab:performance_comparison}}
\centering
\begin{tabular}{|c||c|c|}
\hline
\textbf{Method}          & \textbf{IOU} & \textbf{F1 Score} \\
\hline\hline
HRnet + Concat           & 68.11        & 80.53             \\
\hline
HRnet + Subtract         & 67.62        & 79.81             \\
\hline
HRnet + DAM              & 69.89        & 81.20             \\
\hline
HRnet + SPB + DAM        & 70.42        & 81.63             \\
\hline
\end{tabular}
\end{table}

\begin{table}[h]
\centering
\caption{Ablation Experiment Results of the Detector on the SECD Dataset}
\label{tab:dataset_comparison}
\begin{tabular}{|p{0.25\columnwidth}||c|c|c|c|c|c|}
\hline
\multirow{2}{*}{\textbf{Method}} & \multicolumn{2}{c|}{\textbf{SECD}} & \multicolumn{2}{c|}{\textbf{HRSCD}} & \multicolumn{2}{c|}{\textbf{PSCD-Wuhan}} \\
\cline{2-7}
& \textbf{IOU} & \textbf{F1} & \textbf{IOU} & \textbf{F1} & \textbf{IOU} & \textbf{F1} \\
\hline\hline
Detector+CDCM & 68.63 & 80.75 & 67.88 & 76.75 & 74.24 & 74.57 \\
\hline
\makecell{Detector+CDCM\\+SCM} & 70.42 & 81.63 & 68.07 & 77.17 & 75.26 & 85.51 \\
\hline
\makecell{Detector+CDCM\\+SCM+CLEM} & 70.9 & 82.02 & 68.28 & 77.83 & 77.28 & 86.82 \\
\hline
\makecell{Detector+CDCM\\+SCM\\+CLEM+PLM} & 71.88 & 82.76 & 69.43 & 78.92 & 79.41 & 88.25 \\
\hline
\end{tabular}
\end{table}

\section{Conclusion}
In this letter, we propose the InPhea model, a joint phenology-based change detection model consisting of a detector and a constraint. The detector is responsible for detecting image changes, while the constraint guides the training of the detector by introducing prior knowledge. Within the detector, we construct a Differential Attention Module to enhance the feature representation of change information, supplemented by an HR module and Spatial Pyramid Block to improve detector performance. Within the constraint, we establish four constraint modules to introduce prior information from four perspectives, strengthening the model's ability to filter out phenological pseudo-changes. During model training, we adopt a multi-stage contrastive learning strategy to gradually guide the detector's training. Finally, we conduct comparative experiments and ablation experiments on the HRSCD dataset, SECD dataset, and self-collected dataset PSCD-Wuhan. Experimental results demonstrate the superiority of our InPhea model over other state-of-the-art models, validating the effectiveness of our model in addressing the issue of phenological pseudo-changes.

 

\newpage

\vfill

\end{document}